\documentclass{article}
\usepackage{amsmath,graphicx,mlspconf}
\usepackage{balance}
\usepackage{amssymb,amsfonts,amsthm}
\usepackage[ruled,vlined]{algorithm2e}
\usepackage{enumerate}
\usepackage{textcomp}
\usepackage{rotating}
\usepackage{psfrag}
\usepackage{makecell}
\usepackage{url}
\usepackage[all]{xy}
\usepackage{multirow}
\usepackage{booktabs}
\usepackage[T1]{fontenc}
\usepackage[hidelinks]{hyperref}

\newcommand{\R}{\mathbb{R}}
\newcommand{\C}{\mathbb{C}}
\newcommand{\D}{\mathbb{D}}
\newcommand{\Q}{\mathbb{H}}

\newcommand{\bb}{\begin{equation}}
\newcommand{\ee}{\end{equation}}
\newcommand{\bbb}{\begin{eqnarray}}
\newcommand{\eee}{\end{eqnarray}}
\newcommand{\benu}{\begin{enumerate}}
\newcommand{\eenu}{\end{enumerate}}

\newcommand{\bpm}{\begin{bmatrix}}
\newcommand{\epm}{\end{bmatrix}}

\newcommand{\ii}{\hat{\imath}}
\newcommand{\jj}{\hat{\jmath}}
\newcommand{\kk}{\hat{\kappa}}

\newcommand{\quat}[1]{{#1}_W + {#1}_X \ii + {#1}_Y \jj + {#1}_Z \kk}
\newcommand{\dualq}[1]{\widehat{{#1}}_{W} + \widehat{{#1}}_{X} \ii + \widehat{{#1}}_{Y} \jj + \widehat{{#1}}_{Z} \kk}

\def\boxmax{\kern 0em\hbox{\rm \kern .25em\lower.1ex\hbox{\rlap{$\vee$}}\kern -.07em\lower.2ex\hbox{$\square$}\kern.25em}}
\def\boxmin{\kern 0em\hbox{\rm \kern .25em\lower.1ex\hbox{\rlap{$\wedge$}}\kern -.07em\lower.2ex\hbox{$\square$}\kern.25em}}
\def\boxdiamond{\kern 0em\hbox{\rm \kern .25em\hbox{\rlap{$\diamond$}}\kern -.15em\lower.2ex\hbox{$\square$}}}

\copyrightnotice{979-8-3503-2411-2/23/\$31.00 {\copyright}2023 IEEE}

\toappear{2023 IEEE International Workshop on Machine Learning for Signal Processing, Sept.\ 17--20, 2023, Rome, Italy}

\title{Dual Quaternion Rotational and Translational Equivariance\\ in 3D Rigid Motion Modelling
}
\name{Guilherme Vieira$^*$, Eleonora Grassucci$^\dagger$, Marcos Eduardo Valle$^*$\thanks{This work was partially supported by CNPq under grant no. 315820/2021-7 and FAPESP under grant no. 2022/01831-2.}, Danilo Comminiello$^\dagger$} 
\address{$^*$Dept. of Applied Mathematics, Universidade Estadual de Campinas (Unicamp), Brazil \\$^\dagger$Dept. of Information Engineering, Electronics, and Telecomm., Sapienza University of Rome, Italy}

\begin{document}

\maketitle

\begin{abstract}
Objects' rigid motions in 3D space are described by rotations and translations of a highly-correlated set of points, each with associated $x,y,z$ coordinates that real-valued networks consider as separate entities, losing information. Previous works exploit quaternion algebra and their ability to model rotations in 3D space. However, these algebras do not properly encode translations, leading to sub-optimal performance in 3D learning tasks. To overcome these limitations, we employ a dual quaternion representation of rigid motions in the 3D space that jointly describes rotations and translations of point sets, processing each of the points as a single entity. Our approach is translation and rotation equivariant, so it does not suffer from shifts in the data and better learns object trajectories, as we validate in the experimental evaluations. Models endowed with this formulation outperform previous approaches in a human pose forecasting application, attesting to the effectiveness of the proposed dual quaternion formulation for rigid motions in 3D space.
\end{abstract}

\section{Introduction}
\label{sec:intro}

Human pose forecasting (HPF) aims to predict the position of human keypoints -- such as head, limbs, torso, etc -- in the near future. Accurately estimating the position of human joints and their orientations allows for a better understanding of individuals moving in 3D space. By leveraging this understanding, it is possible to predict the motions of individuals in the immediate future, which is invaluable for applications in self-driving cars \cite{mangalam2020disentangling,razali2021pedestrian}, home healthcare \cite{kidzinski2020deep}, and security surveillance \cite{cormier2022we}, for example. The estimated keypoints can also be used to faithfully reconstruct 3D human body models, having therefore notable applications in augmented and virtual reality \cite{li2022cliff}.

HPF data usually comprise a set of joint data points representing the poses in the space. Such kind of data is inherently a multichannel signal since positions are represented by their coordinates in 3D space. The sequence of positions of a joint in several instants is therefore a time series with highly correlated components. Moreover, joints are part of the same body so they are also correlated among themselves, especially those located in the same limb. Higher dimensional algebras can be used to represent HPF data preserving such correlations.

Learning representations of multichannel data using higher dimension algebras has been proved to outperform real-valued approaches in several problems \cite{talebi15quaternion, grassucci21quaternion, vieira2022acute, Parcollet2020ANetworks}. Quaternion-valued networks already boast works on a significant array of diverse tasks such as audio signal and image processing, computer vision, and dynamic system modeling \cite{Zhou2022TETCI, Jia2022TIP, GrassucciQGAN2021, Brignone2022ISCAS}. 
Human pose estimation using quaternion-valued networks has been done in a few different manners \cite{pavllo2018quaternet,pavllo2020modeling, Hsu2019TMM, Papaioannidis2020TCS}. 
Quaternions are often used since they are able to encapsulate multiple spatial coordinates in one entity and also efficiently represent rotations in 3D space \cite{Zhang2020QuaternionPU}. However, the quaternion product is limited to describing rotations around axes containing the origin. Such operations are not robust to translations, which are extremely common transformations in applications to augmented reality, 3D space modeling, and computer vision.

This work considers an elegant representation for jointly modeling rotations and translations of objects in the 3D space using dual quaternions. Thanks to such representation, we can enclose a full rigid motion, i.e., translation and rotation, in a single entity \cite{pennestri2010dual}, considering object movements in the space as a combination of highly-correlated elements. Differently from previous attempts that only focus on simulated tests \cite{Poppelbaum2022Access, Schwung2021ICIT}, we model human skeleton motions in real-world scenarios, showing the crucial role of the dual quaternion representation in learning body translations in space. Moreover, we provide practical proof that our dual quaternion formulation is both translation and rotation equivariant, which are highly desirable properties for all applications involving 3D movement modeling. Thus, neural models described with our approach significantly mitigate losses from translations and rotations in the data. In the experimental evaluation, we show that dual quaternion models outperform their real and quaternion-valued counterparts according to every metric considered, and visually display the translation and rotation equivariance properties of our approach.

This paper is structured as follows: Section~\ref{sec:quats_and_duals} briefly reviews quaternions and dual numbers. Section~\ref{sec:dual_quats} encompasses the definition of dual quaternions, a description of rigid motions, and the translation and rotation equivariance properties. Section~\ref{sec:Applications} presents the experimental validation, also showcasing the equivariance of dual quaternions. Finally, Section~\ref{sec:concluding-remarks} contains some concluding remarks.

\section{Overview of Quaternions, Dual Numbers, and Dual Quaternions} 
\label{sec:quats_and_duals}

Quaternions, denoted by $\Q$, are one of the most well-known hypercomplex algebras. Commonly seen as an extension to complex numbers $\C$, this algebra was introduced in 1843 by W. R. Hamilton while searching for a domain that could encapsulate operations in three-dimensional space. A quaternion is a number of the form
\bb \label{eq:quat} q = \quat{q} \in \Q, \ee
where $q_W,q_X,q_Y,q_Z$ are real numbers. The symbols $\ii, \jj, \kk$ denote the hyperimaginary units and follow the multiplication rules $\ii^2 = \jj^2 = \kk^2 = \ii\jj\kk = -1$, known as the Hamilton rules. 

A quaternion given by \eqref{eq:quat} efficiently describes a rotation in 3D space. Formally, $q_W$ is called the real part and $\overline{\mathbf{q}} = q_X\ii + q_Y\jj + q_Z\kk$ is referred to as the vector part of $q$. Hence, a quaternion can be written as $q = q_W + \overline{\mathbf{q}}$. We say that $q$ is a pure quaternion if $q_W = 0$. The conjugate of $q = q_W + \overline{\mathbf{q}}$ is easily expressed in this notation as $q^* = q_W - \overline{\mathbf{q}}$. Moreover, we can write a quaternion in polar form
\begin{equation}
\label{eq:quat_polar} 
    q_\theta = \| q \| \left( \cos \left(\frac{\theta}{2}\right) + \sin\left(\frac{\theta}{2}\right) \overline{\mathbf{u}} \right), 
\end{equation}
for $\theta \in [0,2\pi)$ and a pure quaternion $\overline{\mathbf{u}} = u_X\ii + u_Y\jj + u_Z\kk$, where $\| q \| = \sqrt{q_W^2 + q_X^2 + q_Y^2 + q_Z^2}$ is the absolute value of $q$. The rotation of a 3D vector $(p_X,p_Y,p_Z)$ by an angle $\theta$ around the axis determined by $(u_X,u_Y,u_Z)$ can be efficiently determined by
\begin{equation}
\label{eq:quat_rot}
    p_{rot} = qpq^*,    
\end{equation}
where $p = p_X\ii + p_Y\jj + p_Z \kk$ and $u = u_X\ii + u_Y\jj + u_Z\kk$ are pure quaternions encoding the position and axis vectors, and $q$ is the quaternion given by \eqref{eq:quat_polar}.


We would like to remark that, if $q$ is not a unitary quaternion, then $p_{rot}$ given by \eqref{eq:quat_rot} is scaled by $\| q \|^2$. Therefore, it is common to normalize $q$, that is, $q \leftarrow \frac{q}{\| q \|}$. Moreover, the representation in \eqref{eq:quat_rot} is unique due to the domain of $\theta$. Thus, a quaternion uniquely defines a rotation-dilation and vice-versa.

Dual numbers yield a hypercomplex algebra of dimension 2 over $\R$. A dual number has the form $\widehat{a} = a_0 + \varepsilon a_\varepsilon$, where $a_0,a_\varepsilon \in \R$ and $\varepsilon$, called the dual unit, satisfies $\varepsilon^2 = 0$. Using distributivity, we conclude that the product of two dual numbers satisfies
$ (a_0+\varepsilon a_\varepsilon)(b_0+\varepsilon b_\varepsilon) = a_0b_0 + \varepsilon (a_0b_\varepsilon + a_\varepsilon b_0)$.
In particular, $(\varepsilon a_\varepsilon)(\varepsilon b_\varepsilon) = 0$ even if $a_\varepsilon,b_\varepsilon \neq 0$. 

The so-called dual quaternions, denoted by $\D$, are quaternions whose components are dual numbers. In mathematical terms, a dual quaternion is given by $\mathbf{\widehat{q}} = \dualq{q}$, where $\widehat{q}_W,\widehat{q}_X,\widehat{q}_Y,\widehat{q}_Z$ are dual numbers. Interestingly, the dual unit $\varepsilon$ commutes with the quaternion hyperimaginary units $\ii, \jj, \kk$. Thus, the set of dual quaternions $\D$ can be defined equivalently as dual numbers in which each part is a quaternion, i.e., a dual quaternion can be represented by $\mathbf{\widehat{q}} = (q_0 + \varepsilon q_{\varepsilon})$, where $q_0,q_\varepsilon \in \Q$. The norm of a dual quaternion is defined by $\|\mathbf{\widehat{q}}\| = \sqrt{\|q_0\|^2+\|q_\varepsilon\|^2}$ and $\mathbf{\widehat{q}}$ is a unit dual quaternion if $\|\mathbf{\widehat{q}}\|=1$.

In this work, we will 
represent dual quaternions as bold-face letters with a hat $\mathbf{\widehat{q}}$, while dual numbers will be regular letters with hat $\widehat{q}$.

\section{Dual Quaternion Representation of Rigid Motions: Equivariance Properties} \label{sec:dual_quats}



Any rigid motion in 3D space can be reduced to a rotation-translation with respect to a screw axis $\overrightarrow{h}$. Using a proper dual quaternion representation, we can characterize a full rigid motion as a unique entity and leverage algebra properties to build a more robust model \cite{pennestri2010dual}. 

Let us take a unit dual quaternion $\mathbf{\widehat{q}} = \dualq{q}$. On the one hand, the real part $\hat{q}_W$ of $\mathbf{\widehat{q}}$ yields a dual number $\widehat{\theta} = \arccos \left( \widehat{q}_{W} \right) = \frac{\theta}{2} +\varepsilon \frac{s}{2}$,
with $\theta \in [0,2\pi)$. Here, $\frac{\theta}{2}$ is the rotation angle around the screw axis $\overrightarrow{h}$ and $\frac{s}{2}$ is the translation distance along that axis  \cite{pennestri2010dual}. On the other hand, the imaginary part of $\mathbf{\widehat{q}}$ contains information regarding the direction of $\overrightarrow{h}$, which consists of a vector through the origin $\overrightarrow{u}$ translated by a vector $\overrightarrow{d}$. Formally, we have 
$$ \mathbf{\widehat{q}} = \dualq{q}, $$ 
where
\begin{align} \label{eq:dual_quat_polar}
    \begin{split}
        \widehat{q}_{W} &=  \cos \frac{\theta}{2} - \frac{\varepsilon}{2} \overrightarrow{u} \cdot \overrightarrow{d} \sin \frac{\theta}{2} \\
        \widehat{q}_{X} &=  u_x \sin \frac{\theta}{2} + \frac{\varepsilon}{2} \left[ d_x \cos \frac{\theta}{2} - \sin \frac{\theta}{2} (u_yd_z - u_zd_y) \right] \\
        \widehat{q}_{Y} &=  u_y \sin \frac{\theta}{2} + \frac{\varepsilon}{2} \left[ d_y \cos \frac{\theta}{2} - \sin \frac{\theta}{2} (u_zd_x - u_xd_z) \right] \\
        \widehat{q}_{Z} &= u_z \sin \frac{\theta}{2} + \frac{\varepsilon}{2} \left[ d_z \cos \frac{\theta}{2} - \sin \frac{\theta}{2} (u_xd_y - u_yd_x) \right] \\
    \end{split}
\end{align}
Here, $u = u_x \ii + u_y \jj + u_z \kk$ and $d = d_x \ii + d_y \jj + d_z \kk$ are purely imaginary quaternions, and $\overrightarrow{u}\cdot \overrightarrow{d} = u_xd_x + u_yd_y + u_zd_z$ on the first equation denotes the dot product.

Equation \eqref{eq:dual_quat_polar} highlights an interesting property: the dual part is responsible for the translation of the rigid motion. Indeed, if there is no translation the vector representing the translation of the screw axis is $\overrightarrow{d} = 0 \ii + 0 \jj + 0 \kk$. By setting $d_x = d_y = d_z = 0$ in the above equations we find that the dual part becomes null, and the resulting equation is equivalent to \eqref{eq:quat_polar}. This is a straightforward fact: a rigid motion with no translation is simply a rotation around an axis that contains the origin, a motion fully described by a quaternion.
 and 
On the other hand, if we take a rigid motion without rotation, i.e., $\theta = 0$ then \eqref{eq:dual_quat_polar} yields $\mathbf{\widehat{q_t}} = 1 + \frac{\varepsilon}{2} \overrightarrow{d}$. This further reinforces that the dual part is responsible for the translation and shows that the imaginary part of the non-dual part is associated to the rotation operation. This elegant formulation puts in evidence the role played by the dual part and formally shows why dual quaternions can encapsulate both operations while quaternions are restricted to representing rotations. 

Lastly, a unit dual quaternion $\mathbf{\widehat{q}}$ can be written as
\begin{equation} 
\mathbf{\widehat{q}} = \cos \left( \frac{\widehat{\theta}}{2} \right) + \mathbf{\widehat{h}} \sin \left( \frac{\widehat{\theta}}{2} \right),
\end{equation}
where $\mathbf{\widehat{h}}$ is a unit dual quaternion with zero scalar part. Note that this equation is similar to the quaternion polar form \eqref{eq:quat_polar}, except that $\widehat{\theta} = \theta_0 + \varepsilon \theta_\varepsilon$ is a dual number and $\mathbf{\widehat{h}} = {h_0} + \varepsilon h_\varepsilon$ is a dual quaternion. Indeed, this representation yields the screw motion parameters directly: $\theta_0/2$ is the rotation angle around the axis defined by $h_0$ and $\theta_\varepsilon/2$ is the translation along that same axis. $h_\varepsilon$ is the so-called moment of the axis and provides an unambiguous representation of the axis in space. It is defined as $h_\varepsilon = \overrightarrow{p} \times h_0$ where $\overrightarrow{p}$ is a vector from the origin pointing to any point of the axis $h_0$. We note that any choice of vector yields the same moment, since $(\overrightarrow{p} + \alpha h_0) \times h_0 = \overrightarrow{p} \times h_0 = h_\varepsilon$, $\forall \alpha \in \R$. This illustrates one very desirable property: dual quaternions express rigid motions based on an unambiguous representation of the screw axis \cite{kavan2006dual} and thus are independent of the coordinate system, therefore equivariant, as opposed to quaternions whose rotations around axes containing the origin are deformed by translations.

\section{Experimental Evaluation}
\label{sec:Applications}

\subsection{Lorenz System: Equivariance Properties} \label{subsec:lorenz}

The Lorenz system is an ordinary differential equations (ODE) system that describes the movement of a free particle in atmospheric domain effects. This movement is characterized by rigid motions since the particle is under effects of translation at all times with its direction constantly rotating. The equations system is
\begin{equation}
    \left\lbrace \renewcommand\arraystretch{1.5}\begin{array}{l}
     \frac{dx}{dt} =  \sigma (y-x), \\
     \frac{dy}{dt} =  x(\rho - z) -y,\\
     \frac{dz}{dt} = xy - \beta z,
    \end{array} \right.
\end{equation}
where $\sigma, \beta, \rho$ are constants. This system exhibits chaotic behavior for certain values of these constants, such as $\sigma = 10$, $\beta = 8/3$, and $\rho = 28$, meaning that a slight deviation in the initial position results in a large deviation in the particle trajectory. Using these constant values we generate a time series of $10k$ consecutive positions, $10\%$ of which are used for training, and the remaining $90\%$ are used for testing. By strongly limiting the size of the training set we ensure that the performance of the network is more closely related to how well it learns the rigid motions rather than overfitting.





We employ single-hidden-layer MLPs with ReLU activation in the hidden layer and identity in the output layer in this example.
The networks were trained for a one-step-ahead prediction task using a 2-step sliding window, i.e., the network received as inputs two consecutive positions and the desired output is the immediate third one. 
The inputs and outputs for the real-, quaternion-, and dual quaternion-valued models are formatted as follows:
\begin{itemize}
    \item \textbf{Real-Valued Model:} 6 input values containing the 2 positions as $(x_{t-1},y_{t-1},z_{t-1},x_{t},y_{t},z_{t})$, 3 output values as $(x_{t+1},y_{t+1},z_{t+1})$.
    \item \textbf{Quaternion-Valued Model:} 2 quaternion inputs, one for each position, as $(0+x_{t-1}\ii +y_{t-1}\jj +z_{t-1}\kk,0+x_{t}\ii +y_{t}\jj +z_{t}\kk)$, 1 output quaternion as $(0+x_{t+1}\ii +y_{t+1}\jj +z_{t+1}\kk)$.
    \item \textbf{Dual Quaternion-Valued Model:} 1 dual quaternion input containing the 2 positions as $(0+x_{t-1}\ii +y_{t-1}\jj +z_{t-1}\kk) +\varepsilon (0+x_{t}\ii +y_{t}\jj +z_{t}\kk)$, one dual quaternion output $(0+x_{t+1}\ii +y_{t+1}\jj +z_{t+1}\kk) + \varepsilon(0+0\ii +0\jj +0\kk)$.
\end{itemize}
The real network is an MLP with a 6-128-3 architecture, i.e., 6 inputs, 128 neurons in the hidden layer, 3 units in the output layer. Analogously, the quaternion model is an MLP with a 2-80-1 architecture, and the dual quaternion model is an MLP with a 1-53-1 architecture. Networks architectures were chosen to feature an equal number of total trainable parameters, namely 1280. Each model was trained for 250 epochs with the SGD optimizer, a learning rate of $0.0003$ for the real model and $0.0009$ for the other two models, and a single batch consisting of the entire training set.

\begin{figure}[t]
    \centering
    \includegraphics[width=\linewidth]{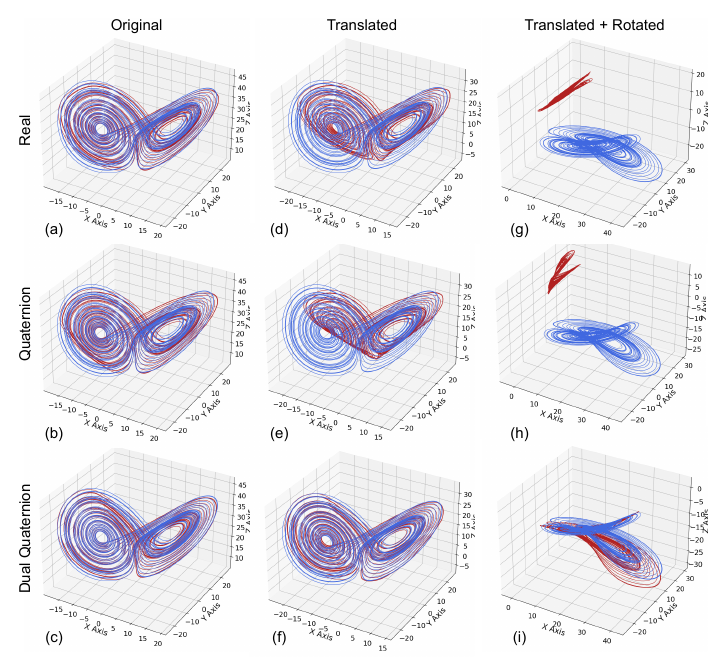}
    \caption{Predicted trajectories in red, expected in blue. Rows contain real-, quaternion- and dual quaternion-valued outputs, respectively. Columns show outputs for the original, translated, and translated+rotated test sets, respectively.}
    \label{fig:lorenz_outs}
\end{figure}

We use the prediction step to showcase the translation and rotation equivariance properties. We generate a random translation and a random rotation. After training on the original training set, the networks are tested on 4 different test sets: i) the original test set; ii) the translated test set; iii) the rotated test set; and iv) the translated+rotated test set. Sets ii-iv) are obtained by applying the aforementioned randomly generated operations.
We reiterate that there is no retraining between the tests. Hence only the original training set was presented to the networks during training. 
Figure \ref{fig:lorenz_outs} shows the predicted trajectories for the original, translated, and translated+rotated test sets.
From the left column, it is clear the models are able to solve the prediction task, albeit with visible flaws from the real and quaternion models (Fig.~\ref{fig:lorenz_outs} \textbf{(a)(b)}). 
In the middle column, it can be seen that the translation barely affects the dual quaternion model, while the other two fail to adapt. Finally, the right column shows the real and quaternion models utterly fail to predict the translated+rotated test set, outputting a trajectory not even close to the expected (Fig.~\ref{fig:lorenz_outs} \textbf{(g)(h)}). On the other hand, the dual quaternion model (Fig.~\ref{fig:lorenz_outs}\textbf{(i)}) shows an accurate prediction, correctly outputting a rotated and translated trajectory. 

Table \ref{tab:exp_lorenz} shows the mean squared error (MSE) and the prediction gain 
per model for each of the 4 test sets. The prediction gain expresses the ratio between a reference level $(\sigma_s)$ and the signal error ($\sigma_e$). Formally, the prediction gain is defined by $R = 10 \log_{10} \frac{\sigma_s^2}{\sigma_e^2}$, 
where $\sigma_s^2$ is the estimated variance of the input signal and $\sigma_e^2$ denotes the estimated variance of the prediction error. It is clear from the table that the dual quaternion model wins every scenario, adapting to the transformations in the test set seamlessly. Since there is no retraining at all, this constitutes strong evidence that the model learns the rigid motions of the Lorenz system rather than mimicking a local behavior. In fact, since a dual quaternion number uniquely describes a rigid motion, the dual quaternion MLP learns the correct rigid motions of the Lorenz system by fine-tuning its dual quaternion parameters. Moreover, this shows that the equivariance properties are a natural consequence of endowing the model with the dual quaternion algebra, which in turn is simply the replacement of sum and product by the respective operations. 

\begin{table*}
\caption{MSE and Prediction Gain for the Lorenz system prediction task with the original, translated, rotated and translated+rotated test sets, respectively.}
\label{tab:exp_lorenz}
\centering
\begin{tabular}{@{}lrrrrrrrr@{}}
\toprule
\multirow{2}{*}{\textbf{Model}}  & \multicolumn{4}{c}{\textbf{Test MSE}\textdownarrow}  & \multicolumn{4}{c}{\textbf{Test Prediction Gain}\textuparrow} \\ 
                        & \textbf{Original} & \textbf{Translated} & \textbf{Rotated} & \textbf{T+R}  
                        & \textbf{Original} & \textbf{Translated}  & \textbf{Rotated} & \textbf{T+R} \\ \midrule
Real-valued & 0.433 & 4.648 & 118.719 & 192.766 & 57.187 & 31.050 & 14.024 & 10.360 \\   
Quaternion-valued  & 0.756 & 7.952 & 178.102 & 263.051 & 51.918 & 24.841 & 6.847 & 3.729 \\
Dual Quaternion-valued  & \textbf{0.272} & \textbf{0.183} & \textbf{2.140} & \textbf{3.617} & \textbf{63.606} &  \textbf{52.722} & \textbf{43.797} & \textbf{40.108} \\
\bottomrule
\end{tabular}
\end{table*}

The possibility to train a network on a dataset and achieve roughly the same performance on translated and/or rotated objects is invaluable and a direct consequence of the rotation and translation equivariance properties of the underlying algebra.


\subsection{Human Pose Forecasting}

We present a variational autoencoder endowed with dual quaternion numbers (DQVAE) that leverages both global and local information by means of dual quaternion representation of rigid motions to more accurately predict the pose of individuals \cite{Parsaeifard2021LearningDR}. Global information is contained in the center of mass coordinates since a human moving is essentially described by a translation applied to each point of the body. Local information, instead, is more accurately described as joint positions relative to the center of mass and more closely related to the action being executed. Indeed, the global trajectory information  is contained in the coordinates $x_c, y_c, z_c$ of the center of mass, while the local information is given by the relative position of joints with respect to the center of mass coordinates as $\left(x-x_c\right), \left(y-y_c\right), \left(z-z_c\right)$.

We consider as baseline the decoupled representations for pose forecasting (DeRPoF) \cite{Parsaeifard2021LearningDR} that involves two different models for trajectory and local pose estimation separately, with an LSTM autoencoder and an LSTM variational autoencoder, respectively. Leveraging the higher-dimensional nature of hypercomplex numbers, we propose the quaternion and dual quaternion coupled representation for pose forecasting models, referred to as Quaternion CoRPoF and Dual Quaternion CoRPoF, respectively. Thanks to the high-dimensional nature of the dual quaternions, the latter model is able to enclose both global and local information in a single entity as
\begin{equation}
    \mathbf{\widehat{q}} = x_c \ii + y_c \jj + z_c \kk + \varepsilon \left( \left(x-x_c\right) \ii + \left(y-y_c\right) \jj + \left(z-z_c\right) \kk \right).
\end{equation}

We carry out a validation experiment in the 3D Poses in the Wild (3DPW) dataset. The task is predicting short-term immediate future poses for human bodies based on joint information. 
The 3DPW dataset contains over $51k$ registered frames from $60$ short videos portraying humans performing basic activities such as hugging, arguing, playing basketball, and dancing, among others. Data is provided in 2 forms: RGB images and structured data. We focus on the latter, which comprises the 3D coordinates of the center of mass and of $13$ key joints of individuals. Hence, this task consists of predicting rigid motions of sets of points, similar to the Lorenz system task above.

The proposed Dual Quaternion CoRPoF has an encoder composed of single-layer dual quaternion LSTM (DQLSTM) blocks of hidden dimension 64, along with two fully connected (DQFC) layers with latent dimension 32, which encode the mean and variance of the latent vector. The decoder of the proposed dual quaternion variational autoencoder (DQVAE) comprises two DQLSTM layers with the same hidden dimension and the final DQFC layers to output the predicted pose. This decoder does not merely reconstruct the input but instead aims at generating future estimations while optimizing the following variational bound:
\begin{equation}
\label{eq:vae_loss}
    \mathcal{L} = \sum_{t=t_{obs}+1}^{t_{obs}+t_{fut}} \| p_t - \hat{p}_t \|^2 + \beta \text{KL} \left( p(z) \| \mathcal{N} (\mathbf{0}, \mathbf{I}) \right),
\end{equation}
where $p_t$ is the future ground truth pose, $\hat{p}_t$ is the predicted one, $p(z)$ is the latent distribution, and $\beta$ is the hyperparameter that balances the two elements. The first term of \eqref{eq:vae_loss} cares about the trajectory estimation, while the second element is crucial for local poses. The model is trained for $1000$ epochs by the Adam optimizer with a learning rate of $0.01$ and an adaptive scheduler, as suggested in \cite{Parsaeifard2021LearningDR}. It is worth mentioning that due to the properties of the Hamilton product, the quaternion and dual quaternion models reduce the number of parameters of the network with respect to the real-valued baseline CoRPoF by $75\%$ and $88\%$, respectively.

\begin{figure}[t]
    \centering
    \includegraphics[width=0.45\textwidth]{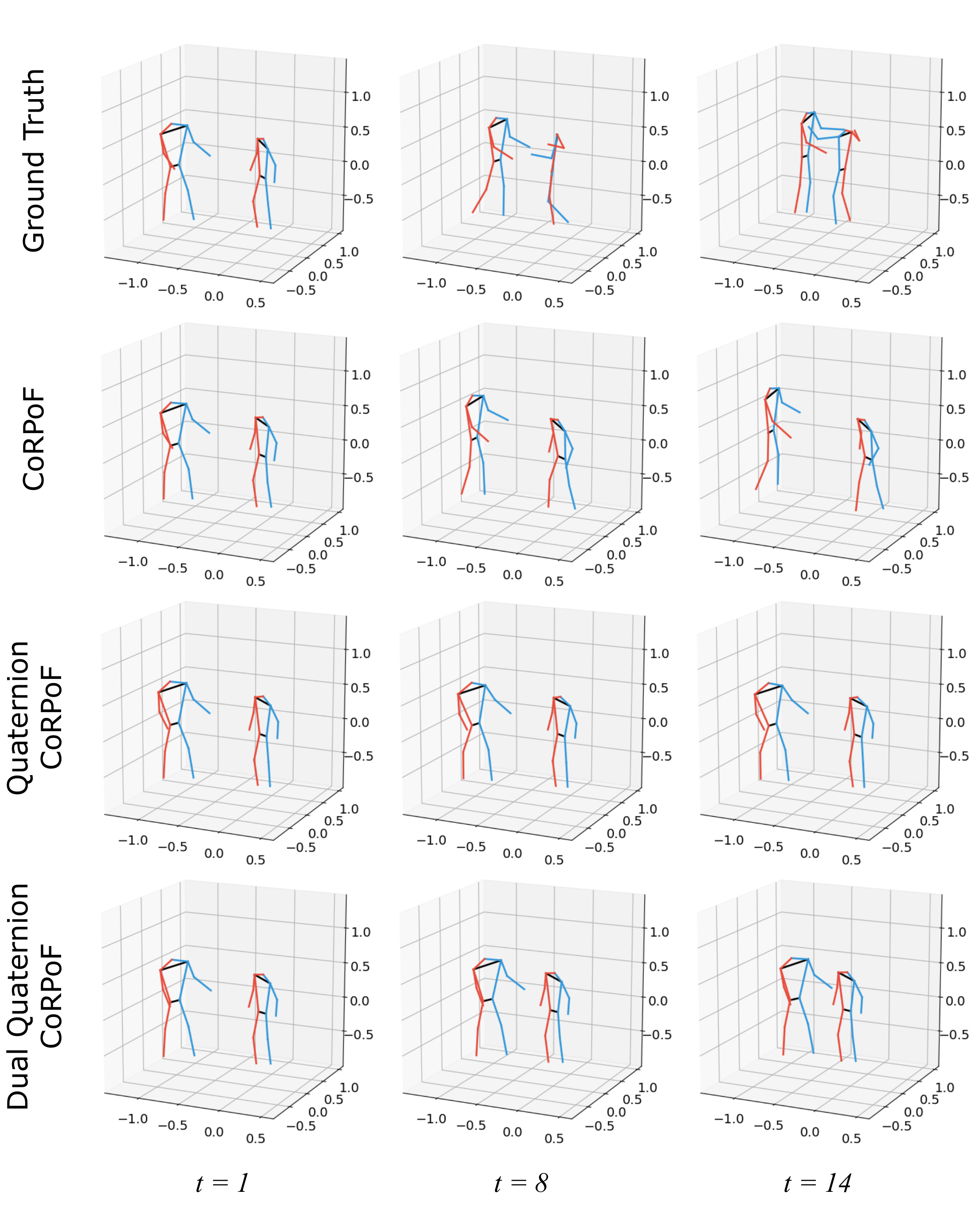}
    \caption{Estimated poses at different time steps. The dual quaternion CoRPoF models the translations, while the real- and quaternion-valued networks fail to do so.}
    \label{fig:3dpw_skeletons}
\end{figure}

\begin{table}
\caption{Results for 3D human pose estimation on the 3DPW dataset. VIM scores are in centimeters, as suggested in \cite{Parsaeifard2021LearningDR}.}
\label{tab:res}
\centering
\resizebox{\linewidth}{!}{
\begin{tabular}{@{}lcccc@{}}
\toprule
\textbf{Model} & \textbf{VIM}$\downarrow$ & \textbf{FDE}$\downarrow$ & \textbf{Val Loss}$\downarrow$ \\ \midrule
SC-MPF \cite{Adeli2020SociallyAC} & 46.28 & - & - \\   
Nearest Neighbour \cite{Zhang2019Predicting3H} & 27.34 & - & - \\
Zero Velocity \cite{Martinez2017OnHM} & 29.35 & - & - \\
DeRPoF \cite{Parsaeifard2021LearningDR} & 19.07 \scriptsize $\pm$ .005 & 0.360 \scriptsize $\pm$ .007 & -\\
CoRPoF \cite{Parsaeifard2021LearningDR} & 16.76 \scriptsize $\pm$ .003 & 0.317 \scriptsize $\pm$ .001 & 0.118 \scriptsize $\pm$ .004\\
Quaternion CoRPoF & 16.35 \scriptsize $\pm$ .009 & 0.271 \scriptsize $\pm$ .002 & 0.105 \scriptsize $\pm$ .010 \\
Dual Quaternion CoRPoF & \textbf{15.23} \scriptsize $\pm$ .002 & \textbf{0.266} \scriptsize $\pm$ .001 & \textbf{0.103} \scriptsize $\pm$ .006 \\ \bottomrule
\end{tabular}}
\end{table}

To assess the performance of our model and to be consistent with the previous literature we compute the visibility ignored metric (VIM), which is the average of the distances between each predicted joint and the corresponding point in the ground truth, and the final displacement error (FDE), which is instead an L2 distance. Table~\ref{tab:res} reports these objective metrics as average scores and standard deviations over multiple runs. As expected, there is notable improvement operated by our dual quaternion formulation due to the effectiveness of higher-dimensional representations for movements in the 3D space. The comparisons with SC-MPF, Nearest Neighbours, and Zero velocity are performed with the scores reported in their papers, as their code is not available. Figure~\ref{fig:3dpw_skeletons} displays a ground truth sample and the corresponding predicted skeletons by the Dual Quaternion CoRPoF and its quaternion- and real-valued counterparts. Although real CoRPoF tries to model local poses, it fails to learn the right movement trajectory and skeletons are stuck in the same coordinates. The quaternion model on the other hand shows little translation of the skeletons but fails to model their fine movements. In the last row, the dual quaternion CoRPoF learns the correct trajectory of both skeletons while also displaying some of the expected fine movements. This showcases the natural ability of the proposed dual representation to model translations in 3D space.

\section{Concluding Remarks} \label{sec:concluding-remarks}
In this paper, we present a formulation of dual quaternions that puts in evidence its connection to modeling rigid motions in 3D. We formally explain how a dual quaternion jointly encases information regarding translations and rotations, and provide a practical example of the translation and rotation equivariance properties using the Lorenz system. We proceed to show how models endowed with this formulation outperform other approaches in an application to human pose forecasting, proving our theoretical claims that the dual quaternion-valued models are robust to data shifting and more adequate to such tasks.


\ninept
\bibliographystyle{IEEEbib}
\bibliography{references.bib}

\begin{thebibliography}{10}

\bibitem{mangalam2020disentangling}
K.~Mangalam, E.~Adeli, K.-H. Lee, A.~Gaidon, and J.~C. Niebles,
\newblock ``Disentangling human dynamics for pedestrian locomotion forecasting
  with noisy supervision,''
\newblock in {\em IEEE/CVF Winter Conference on Applications of Computer Vision
  (WCACV)}, 2020, pp. 2784--2793.

\bibitem{razali2021pedestrian}
H.~Razali, T.~Mordan, and A.~Alahi,
\newblock ``Pedestrian intention prediction: A convolutional bottom-up
  multi-task approach,''
\newblock {\em Transportation research part C: emerging technologies}, vol.
  130, pp. 103259, 2021.

\bibitem{kidzinski2020deep}
L.~Kidzi{\'n}ski, B.~Yang, J.~L. Hicks, A.~Rajagopal, S.~L. Delp, and M.~H.
  Schwartz,
\newblock ``Deep neural networks enable quantitative movement analysis using
  single-camera videos,''
\newblock {\em Nature communications}, vol. 11, no. 1, pp. 1--10, 2020.

\bibitem{cormier2022we}
M.~Cormier, A.~Clepe, A.~Specker, and J.~Beyerer,
\newblock ``Where are we with human pose estimation in real-world
  surveillance?,''
\newblock in {\em IEEE/CVF Winter Conference on Applications of Computer Vision
  (WCACV)}, 2022, pp. 591--601.

\bibitem{li2022cliff}
Z.~Li, J.~Liu, Z.~Zhang, S.~Xu, and Y.~Yan,
\newblock ``{CLIFF}: Carrying location information in full frames into human
  pose and shape estimation,''
\newblock {\em arXiv preprint arXiv:2208.00571}, 2022.

\bibitem{talebi15quaternion}
S.~P. Talebi and D.~P. Mandic,
\newblock ``A quaternion frequency estimator for three-phase power systems,''
\newblock in {\em {IEEE} Int. Conf. on Acoustics, Speech and Signal Process.
  ({ICASSP})}, 2015, pp. 3956--3960.

\bibitem{grassucci21quaternion}
E.~Grassucci, D.~Comminiello, and A.~Uncini,
\newblock ``A quaternion-valued variational autoencoder,''
\newblock in {\em {EEE} Int. Conf. on Acoustics, Speech and Signal Process.
  ({ICASSP})}, 2021, pp. 3310--3314.

\bibitem{vieira2022acute}
G.~Vieira and M.~E. Valle,
\newblock ``Acute lymphoblastic leukemia detection using hypercomplex-valued
  convolutional neural networks,''
\newblock {\em arXiv preprint arXiv:2205.13273}, 2022.

\bibitem{Parcollet2020ANetworks}
T.~Parcollet, M.~Morchid, and G.~Linar{\`{e}}s,
\newblock ``{A survey of quaternion neural networks},''
\newblock {\em Artificial Intelligence Review}, vol. 53, no. 4, pp. 2957--2982,
  4 2020.

\bibitem{Zhou2022TETCI}
Y.~Zhou, L.~Jin, G.~Ma, and X.~Xu,
\newblock ``Quaternion capsule neural network with region attention for facial
  expression recognition in color images,''
\newblock {\em IEEE Transactions on Emerging Topics in Computational
  Intelligence}, vol. 6, no. 4, pp. 893--912, 2022.

\bibitem{Jia2022TIP}
Z.~Jia, Q.~Jin, M.~K. Ng, and X.-L. Zhao,
\newblock ``Non-local robust quaternion matrix completion for large-scale color
  image and video inpainting,''
\newblock {\em IEEE Trans. on Image Process.}, vol. 31, pp. 3868--3883, 2022.

\bibitem{GrassucciQGAN2021}
E.~Grassucci, E.~Cicero, and D.~Comminiello,
\newblock ``Quaternion generative adversarial networks,''
\newblock in {\em Generative Adversarial Learning: Architectures and
  Applications}, R.~Razavi-Far, A.~Ruiz-Garcia, V.~Palade, and J.~Schmidhuber,
  Eds., pp. 57--86. Springer International Publishing, Cham, 2022.

\bibitem{Brignone2022ISCAS}
C.~Brignone, G.~Mancini, E.~Grassucci, A.~Uncini, and D.~Comminiello,
\newblock ``Efficient sound event localization and detection in the quaternion
  domain,''
\newblock {\em IEEE Trans. on Circuits and Systems II: Express Brief}, vol. 69,
  no. 5, pp. 2453--2457, 2022.

\bibitem{pavllo2018quaternet}
D.~Pavllo, D.~Grangier, and M.~Auli,
\newblock ``Quaternet: A quaternion-based recurrent model for human motion,''
\newblock {\em arXiv preprint arXiv:1805.06485}, 2018.

\bibitem{pavllo2020modeling}
D.~Pavllo, C.~Feichtenhofer, M.~Auli, and D.~Grangier,
\newblock ``Modeling human motion with quaternion-based neural networks,''
\newblock {\em Int. Journal of Computer Vision}, vol. 128, no. 4, pp. 855--872,
  2020.

\bibitem{Hsu2019TMM}
H.-W. Hsu, T.-Y. Wu, S.~Wan, W.~H. Wong, and C.-Y. Lee,
\newblock ``{QuatNet}: Quaternion-based head pose estimation with
  multiregression loss,''
\newblock {\em IEEE Trans. on Multimedia}, vol. 21, no. 4, pp. 1035--1046,
  2019.

\bibitem{Papaioannidis2020TCS}
C.~Papaioannidis and I.~Pitas,
\newblock ``{3D} object pose estimation using multi-objective quaternion
  learning,''
\newblock {\em {IEEE} Trans. on Circuits and Systems for Video Technology},
  vol. 30, no. 8, pp. 2683--2693, 2020.

\bibitem{Zhang2020QuaternionPU}
X.~Zhang, S.~Qin, Y.~Xu, and H.~Xu,
\newblock ``Quaternion product units for deep learning on 3d rotation groups,''
\newblock {\em {IEEE/CVF} Conf. on Computer Vision and Pattern Recognition
  ({CVPR})}, pp. 7302--7311, 2020.

\bibitem{pennestri2010dual}
E.~Pennestr{\`\i} and P.~P. Valentini,
\newblock ``Dual quaternions as a tool for rigid body motion analysis: A
  tutorial with an application to biomechanics,''
\newblock {\em Archive of Mechanical Engineering}, vol. 57, no. 2, pp.
  187--205, 2010.

\bibitem{Poppelbaum2022Access}
J.~Pöppelbaum and A.~Schwung,
\newblock ``Predicting rigid body dynamics using dual quaternion recurrent
  neural networks with quaternion attention,''
\newblock {\em IEEE Access}, vol. 10, pp. 82923--82943, 2022.

\bibitem{Schwung2021ICIT}
A.~Schwung, J.~Pöppelbaum, and P.~C. Nutakki,
\newblock ``Rigid body movement prediction using dual quaternion recurrent
  neural networks,''
\newblock in {\em {IEEE} Int. Conf. on Industrial Technology ({ICIT})}, 2021,
  vol.~1, pp. 756--761.

\bibitem{kavan2006dual}
L.~Kavan, S.~Collins, C.~O’Sullivan, and J.~Zara,
\newblock ``Dual quaternions for rigid transformation blending,''
\newblock {\em Trinity College Dublin, Tech. Rep.}, vol. 46, 2006.

\bibitem{Parsaeifard2021LearningDR}
B.~Parsaeifard, S.~Saadatnejad, Y.~Liu, T.~Mordan, and A.~Alahi,
\newblock ``Learning decoupled representations for human pose forecasting,''
\newblock {\em {IEEE/CVF} Int. Conf. on Computer Vision Workshops ({ICCVW})},
  pp. 2294--2303, 2021.

\bibitem{Adeli2020SociallyAC}
V.~Adeli, E.~Adeli, I.~D. Reid, J.~C. Niebles, and H.~Rezatofighi,
\newblock ``Socially and contextually aware human motion and pose
  forecasting,''
\newblock {\em {IEEE} Robotics and Automation Letters}, vol. 5, pp. 6033--6040,
  2020.

\bibitem{Zhang2019Predicting3H}
J.~Y. Zhang, P.~Felsen, A.~Kanazawa, and J.~Malik,
\newblock ``Predicting 3d human dynamics from video,''
\newblock {\em {IEEE/CVF} Int. Conf. on Computer Vision ({ICCV})}, pp.
  7113--7122, 2019.

\bibitem{Martinez2017OnHM}
J.~Martinez, M.~J. Black, and J.~Romero,
\newblock ``On human motion prediction using recurrent neural networks,''
\newblock {\em {IEEE} Conf. on Computer Vision and Pattern Recognition
  ({CVPR})}, pp. 4674--4683, 2017.

\end{thebibliography}

\end{document}